\documentclass[conference]{IEEEtran}
\IEEEoverridecommandlockouts

\usepackage{cite}
\usepackage{amsmath,amssymb,amsfonts}
\usepackage{algorithmic}
\usepackage{graphicx}
\usepackage{textcomp}
\usepackage{xcolor}
\usepackage{booktabs}
\usepackage{multirow}
\usepackage{url}
\usepackage[hang,flushmargin]{footmisc}
\def\BibTeX{{\rm B\kern-.05em{\sc i\kern-.025em b}\kern-.08em
    T\kern-.1667em\lower.7ex\hbox{E}\kern-.125emX}}

\usepackage{subcaption}
\usepackage{url}

\newcommand{\ieno}{\textit{i}.\textit{e}.} 
\newcommand{\egno}{\textit{e}.\textit{g}.}

\newcommand{\myparagraph}[1]{{\vspace{.2em} \noindent \bf #1}}

\newcommand{\Rmnum}[1]{\expandafter@slowromancap\romannumeral #1@}
\begin{document}
%
\title{Towards Defining an Efficient and Expandable \\ File Format for AI-Generated Contents \\


\author{\IEEEauthorblockN{Yixin Gao$^\dagger$, Runsen Feng$^\dagger$,Xin Li$^*$, Weiping Li, Zhibo Chen$^*$}
\IEEEauthorblockA{
\textit{University of Science and Technology of China} \\
\{gaoyixin, fengruns\}@mail.ustc.edu.cn, \{xin.li, wpli, chenzhibo\}@ustc.edu.cn
}}
\thanks{$^\dagger$Yixin Gao and Runsen Feng contribute equally to this work. \\ $^*$Corresponding authors: Xin Li (xin.li@ustc.edu.cn) and Zhibo Chen (chenzhibo@ustc.edu.cn). This work was supported in part by NSFC under Grant 623B2098, 62371434, and 62021001.}}

\maketitle

\begin{abstract}
Recently, AI-generated content (AIGC) has gained significant traction due to its powerful creation capability. However, the storage and transmission of large amounts of high-quality AIGC images inevitably pose new challenges for recent file formats. To overcome this, we define a new file format for AIGC images, named AIGIF, enabling ultra-low bitrate coding of AIGC images. Unlike compressing AIGC images intuitively with pixel-wise space as existing file formats, AIGIF instead compresses the generation syntax. This raises a crucial question: Which generation syntax elements, e.g., text prompt, device configuration, etc, are necessary for compression/transmission? To answer this question, we systematically investigate the effects of three essential factors: platform, generative model, and data configuration. We experimentally find that a well-designed composable bitstream structure incorporating the above three factors can achieve an impressive compression ratio of even up to 1/10,000 while still ensuring high fidelity. We also introduce an expandable syntax in AIGIF to support the extension of the most advanced generation models to be developed in the future.

\end{abstract}

\IEEEpeerreviewmaketitle

\section{Introduction}
\label{sec:intro}
In recent years, artificial intelligence-generated content (AIGC) has garnered significant attention and experienced remarkable advancements since its powerful interactive and creation capability. Particularly, 
text-to-image generation techniques~\cite{rombach2022high, Imagen, Midjourney, DiT,betker2023improving,dai2023emu} have been widely used for various high-quality image creation, which enables users to create multiple high-fidelity and diverse content based on input text prompts in a short time, including portraits, landscapes, abstract art, and even images crafted in the styles of well-known artists. On the popular AIGC sharing platform Civitai, the number of newly created images can reach up to 2.4 million in just one week\footnote{\url{https://civitai.com/articles/5278/whats-new-this-week-with-civitai-5102024}}. However, the massive production of AIGC poses significant challenges for high-quality data storage and transmission with the existing image file format, \egno, PNG.

\begin{figure}[htb]
 \centering
 \begin{subfigure}{0.9\linewidth}
\includegraphics[width=\linewidth, clip, trim=0cm 0cm 0cm 0cm]{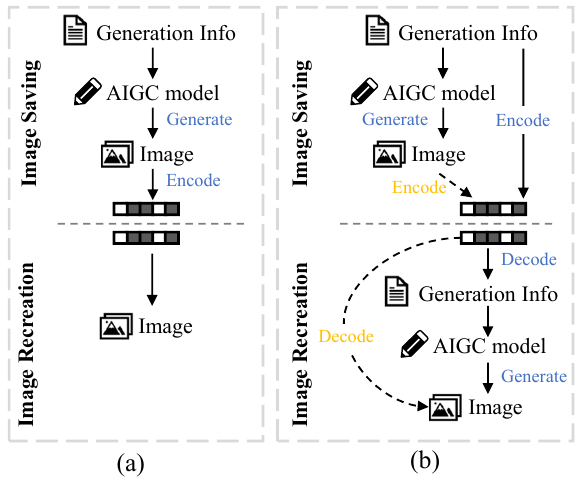}
 \end{subfigure}

 \caption{High-level comparison of image saving and image recreation process between (a) common image formats like PNG~\cite{png} and (b) our proposed image format. Rather than directly saving compressed image pixels into a file, we save the compact generation information into a file as the representation of AI-generated images. Dash line is optional.}
\label{fig:pipeline}
\end{figure}
\begin{figure*}[t] 
\centering
\includegraphics[width=0.75\textwidth]{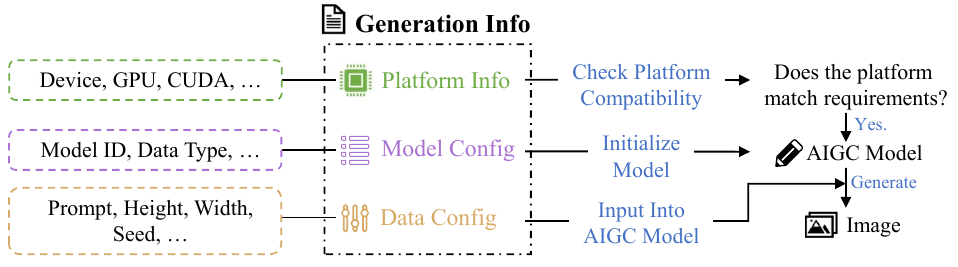}
\caption{Overview of the generation information for reproducing the image generation process.}
\label{fig:overview}
\end{figure*}

Notably, most commonly used image file formats like PNG~\cite{png} and JPEG-XL~\cite{jpegxl_1, jpegxl_2} compress an image into a file by performing the transform and entropy coding to reduce the spatial redundancy. Despite that, image compression in the pixel-wise space merely achieves the limited compression ratio of approximately 1/2 to 1/12 for high-fidelity image compression~\cite{rahman2022comparative,bai2024deep}, which implies that it still requires hundreds of kilobytes or even several megabytes to store a high-resolution AI-generated image. To overcome the above challenge, we aim to break through the limitation of existing image file formats for AIGC images, achieving ultra-low bitrate compression while maintaining the high fidelity of AIGC images. 


In this work, we define a new efficient and expandable \textbf{AI}-\textbf{G}enerated \textbf{I}mage \textbf{F}ormat, named AIGIF, which is capable of achieving high-fidelity AIGC image compression with an ultra-low bitrate. Notably, unlike natural images, AIGC images are generated with generative models (\egno, Stable Diffusion v1.5) by user-defined data configuration like text prompts and size. Since the large computational cost, a particular platform (\egno, CPU or GPU) is required to support the generation process. Meanwhile, the creation process and generation syntax of AIGC images are usually accessible for users, which raises an intuitive question for us: ``whether we can compress AIGC images by compressing generation syntax only instead of images themselves like existing image file formats".

Inspired by this, we propose a composable bitstream structure for our AIGIF with three essential generation syntax elements in the text-to-image generation process: platform, generation model, and data configurations. Concretely, we systematically experimentally investigate the effects of the above three factors on AIGC image reconstruction. We can obtain two interesting and crucial experimental conclusions: i) data and model configurations determine the content consistency of AIGC images; ii) the change of platform configuration will cause slight degradation for AIGC images due to the differences of floating-point arithmetic across different devices. Considering that, we achieve the AIGIF by compressing the above three factors, \ieno, the generation syntax elements, with lossless coding. The whole pipeline of AIGC coding can be found in Fig.~\ref{fig:pipeline}. To cope with the loss of generation syntax elements in some special cases, our AIGIF also supports the direct compression and transmission of original AIGC images (shown as the dash lines in Fig.~\ref{fig:pipeline}). Considering the rapidly evolved AIGC techniques, we introduce an expandable syntax, termed ``exp code" in our AIGIF, which employs a special byte value ``11111111" in bitstream to request the reallocation of an extra byte to store the index of newly developed generative models, thereby enabling the powerful applicability of AIGIF.
 With a well-designed composable bitstream structure, our AIGIF could achieve ultra-low-bitrate compression for AIGC images with a compression ratio of up to 1/10,000, while still ensuring high fidelity. 
 
\section{Methodology}
\label{sec:metho}
\subsection{Preliminary}
\myparagraph{Image File Formats} aim to reduce the image size for storage and transmission by compressing images into bitstream. Meanwhile, it also ensures the standardization of the image encoding and decoding process. Representative image file formats can be roughly divided into lossless image formats, such as PNG~\cite{png}, and lossy image formats, such as JPEG~\cite{jpeg2000} and WebP~\cite{webp}, where lossless image format is usually applied to meet the high-fidelity requirements. In this paper, we select two representative high-fidelity image formats, including PNG~\cite{png} and JPEG-XL~\cite{jpegxl_2,jpegxl_1} for comparison and analysis.

\myparagraph{AIGC} has paved the path to general artificial intelligence(AGI) by empowering AI models with the powerful creation, understanding, and interactive capabilities. 
As the representative task of AIGC, text-to-image generation has progressed significantly with the advancement of diffusion models~\cite{sohl2015deep, ho2020denoising, song2020score,li2023diffusion} and multi-modality methods like CLIP~\cite{radford2021learning}. In this work, we adopt the popular Stable Diffusion~\cite{rombach2022high} to produce the AIGC images. Given the fine-grained description prompts, users can generate amounts of high-quality images they want, which can assist the artistic creation or entertainment, posing a significant challenge to existing image file formats in terms of storage and transmission.

\subsection{AIGIF}
Although existing image formats have been greatly developed, high-fidelity compression of AIGC images can only achieve a compression ratio of about 1/2 to 1/12. This means general pixel-wise space compression, including transformation, and entropy coding, is not able to support high-fidelity and ultra-low-bitrate compression simultaneously. To overcome this, we define an innovative efficient, and expandable image format, named AIGIF, which takes the first step to compress the generation information of AIGC images instead of their pixels. In the subsections, we will clarify the details for AIGIF as: 1) observations, 2) Composable Bitstream Structure, 3) Expandable Syntax, and 4) Overall Pipeline.

\subsubsection{Observations}
The first step is to identify which generation syntax elements are crucial for our AIGIF. Therefore, we systematically investigate the effect of platform, generative model, and data configuration with analysis and well-designed experiments. 
The input data configuration decides the attribute of the generated image. For example, a prompt is a sentence serving as the input text conditions of AIGC models, which is semantically related to the generated content. Height and width determine the spatial resolution of the image, and seed controls the pseudo-random generator for noise sampling in generative models. Any absence of these elements will drastically change the image composition. 
From Table~\ref{tab:sdv1.5}, we can observe that the inconsistency of running platforms slightly degrades the recreated AIGC images. 
As shown in Fig~\ref{fig:crossmodel}, even with the same data and platform configuration, different AIGC models generate completely inconsistent images. These observations inspired us to achieve a composable bitstream structure with the above three essential elements. 

\subsubsection{Composable Bitstream Structure}
As shown in Table~\ref{tab:file_format}, we provide the descriptions, data type and example values of the file format example for text-to-image generative models.
From top to bottom, there are four components: compression options, platform, generative model, and data configurations. 
Among them, compression options are irrelevant to the generation process. Specifically, "saving pixels" determines if the original pixels need to be compressed into a file, ensuring recreation when users cannot reproduce the generation process. The "pixel compressor" determines the pixel compression method if needed, while the "text compressor" specifies the method used to compress any string-type information. The "model compressor" denotes the neural network compression methods for AIGC models.
The second part is platform configuration, specifying the hardware and software environments used during the image generation process. 
Next, the generative model configuration identifies the specific AIGC model and its relevant settings used to generate the image. At last, the data configuration includes all the parameters that define the attributes of the generated image. 
Besides the generation information described previously, we here also provide the model-dependent information within ``()'' in the table.

\subsubsection{Expandable Syntax}
Given the rapid evolution of AIGC techniques, it is essential to support the expansion of new AIGC models. To address this, we introduce an expandable syntax for model IDs, termed "exp code" in our AIGIF. The "exp code" starts as a 1-byte syntax and uses a special value (e.g., 11111111) to signal the need for additional bytes. Each added byte can support up to 254 new entries, with the last entry being the expandable code again. This recursive structure ensures an adaptable and scalable model identification system.

\subsubsection{Overall Pipeline}
Fig~\ref{fig:pipeline} illustrates the pipeline of our AIGIF. Specifically, in the image-saving process, when an image is created by a user, the generation syntax of the corresponding generation process is recorded. The generation information is then losslessly compressed into a file. 
For image recreation, as shown in Fig.~\ref{fig:overview}, a user first entropy decodes the generation information from an AIGIF file. The decoded platform information is used to check platform compatibility to ensure cross-platform recreation. If the platform used for image recreation is compatible with that of the generation process, the recreation process will initialize the model by the model configuration. Subsequently, the data configuration is inputted into the initialized AIGC model for image recreation.

Additionally, to handle the loss of generation syntax elements in special cases, our AIGIF supports direct compression and transmission of original AIGC images (shown as dashed lines in Fig. 1)~\cite{png,jpegxl_1,wu2021learned,guo2021causal,feng2023nvtc}. This feature ensures high-fidelity image preservation and transmission even when the generation syntax cannot be fully utilized. 
By employing a well-defined composable bitstream structure and offering flexible options for image compression and recreation, AIGIF provides a solution for efficiently reproducing high-quality AI-generated images.

\begin{table}[htb]
    \centering
    \caption{A file format example for text-to-image generative models. The descriptions within ``()'' depend on the model ID, where different model IDs may have different descriptions. The ``exp code'' means expandable code, which uses 1 byte as the basic unit and employs a special value (\egno, 1111111) to signal the allocation of an additional byte for identifying new models, allowing for flexible and endless expansion.}
    \resizebox{0.5\textwidth}{!}{
    \begin{tabular}{cccc}
    \toprule
        Description & Type &  Example Value \\
        \midrule
        saving pixels &  1 bit &  0=No; 1=Yes\\
        pixel compressor  &  4 bits &  0=None; 1=png \\
        text compressor &  4 bits &  0=None; 1=zlib \\
        saving model &  1 bit &  0=No; 1=Yes\\
        model compressor &  4 bits &  0=None; 1=int8 \\
        \hline
        device &  4 bits &  0=``CPU''; 1=``GPU''\\
        gpu  &  1 byte &  1=``NVIDIAGeForceGTX1080Ti'' \\
        cuda & 1 byte &  1=``cu121'' \\
        ... & Any &  ... \\
        \hline
        model ID & exp code &  0=``stable-diffusion-v1-5'' \\
        data type & 4 bits &  0=``float32''; 1=``float16'' \\
        (scheduler)  & 4 bits & 0=DDIM \\
        ... & Any & ... & \\
        \hline
        prompt & String &  ``A cute cat'' \\
        negative prompt & String &  ``worst quality'' \\
        height & 4 bytes &  1024   \\
        width & 4 bytes &  1024 \\
        seed  & 4 bytes &  829557441 \\
        (diffusion steps) & 2 bytes &  25 \\
        (guidance scale) & 4 bytes &  7.5 \\
        ... & Any &  ... \\
    \bottomrule
\end{tabular}}
\label{tab:file_format}
\end{table}



\section{Experiments}
\label{sec:exp}

\subsection{Experimental Setup}
To evaluate the file size of different image formats on AI-generation images, we make a dataset that contains 120 images generated by 3 popular text-to-image generative models: Stable-Diffusion (SD) 1.5, SD2.1~\cite{rombach2022high} and SDXL~\cite{sdxl}. We employ each model to generate 40 images using a DDIM sampler~\cite{ddim} with the following hyperparameters: number of diffusion steps $T$ = 50, guidance scale $w$ = 7.5, height $h$ = 1024 and width $w$ = 1024. We use Lempel-Ziv coding as implemented in the zlib library~\cite{zlib} compressor to compress the configurations and conditions in the text-to-image generation process and evaluate the complexity on an RTX 3090 GPU.


\subsection{Results}
\vspace{-2mm}
\begin{table}[h]
    \centering
    \caption{Comparisons of different image file formats on AIGC images.}
    \vspace{-1mm}
    \begin{tabular}{ccccc}
    \toprule
         Methods& Pixel Data &  PNG & JPEG-XL & AIGIF (Ours)\\
         \midrule
         Rate (byte) &  3,145,728 &  1,907,472 & 1,291,602 & 215 \\
         Dec Time (s) &  - &  0.03 & 0.40 & 48.46 \\
    \bottomrule
    \end{tabular}
    \label{tab:lossless_coding_performance}
    \vspace{-2mm}
\end{table}
\myparagraph{File size comparison.} Table \ref{tab:lossless_coding_performance} presents a comparison between our AIGIF file format and current state-of-the-art (SOTA) file formats in lossless image compression, measured in bytes per image. By compressing the generation information of the text-to-image process rather than directly removing redundancy between pixels, AIGIF achieves a compression ratio of over 10,000 times compared to pixel data. 
Furthermore, AIGIF significantly outperforms the widely-used PNG~\cite{png} and the SOTA lossless image file format JPEG-XL~\cite{jpegxl_1,jpegxl_2}, with an average file size of just 215 bytes, far less than the over 1 MB required by the other methods.
\begin{table}[h]
\centering
\caption{Image saving and recreation time complexity. Generation denotes the average image generation time.}
\vspace{-1mm}
\resizebox{0.45\textwidth}{!}{
\begin{tabular}{@{}c|cc|cc@{}}
\toprule
Model & \multicolumn{2}{c|}{Image Saving (s)} & \multicolumn{2}{c}{Image Recreation (s)} \\ \midrule
     & Generation & Entropy Coding & Generation & Entropy Decoding \\ \midrule
SD1.5 & 48.4622 & 0.0003 & 48.4589 & 0.0003 \\ 
Hyper-SD & 1.1132 & 0.0003 & 1.1174 & 0.0003 \\ \bottomrule
\end{tabular}}
\vspace{-5mm}
\label{tab:complexity}
\end{table}

\myparagraph{Complexity.} We present decoding time comparisons in Table~\ref{tab:lossless_coding_performance} and a detailed complexity analysis in Table~\ref{tab:complexity}.
As shown in Table~\ref{tab:complexity}, the time complexity of our AIGIF comprises two main components: the image generation process and the entropy coding of generation information. Notably, AIGIF involves a user-defined generative process, making the total time complexity highly dependent on the image generation time. Although the standard SD1.5 model typically takes around 50 seconds to generate a 1024$\times$1024 image, considerable research has focused on accelerating the diffusion process. Techniques such as distillation~\cite{salimans2022progressive,meng2023distillation,hypersd}, consistency models~\cite{song2023consistency,latentconsistency}, and faster samplers~\cite{dpm,zheng2024dpm} have significantly reduced the required diffusion steps. These acceleration techniques are orthogonal to our approach. Here we provide the time complexity of two models, SD1.5 and Hyper-SD, where Hyper-SD is a SOTA diffusion model acceleration technique~\cite{hypersd}. As demonstrated in Table~\ref{tab:complexity}, the 1-step Hyper-SD method only requires about one second for image generation, significantly reducing decoding complexity.

\vspace{-2mm}
\begin{table}[htb]
    \centering
    \caption{Notations of different CPU and GPU devices.}
    \vspace{-2mm}
    \resizebox{0.45\textwidth}{!}{
    \begin{tabular}{cc}
    \toprule
        Notation & Device \\
        \midrule
         CPU~1 & Intel(R) Xeon(R) Gold 6248R CPU @ 3.00GHz \\
         CPU~2 & 13th Gen Intel(R) Core(TM) i7-13700F \\
         CPU~3 & Intel(R) Xeon(R) CPU E5-2699 v4 @ 2.20GHz \\
         GPU~1 \& GPU~2& NVIDIA RTX 3090 \\
         GPU~3& NVIDIA GTX 1080ti \\
    \bottomrule
    \end{tabular}
    }
    \label{tab:notation}
\end{table}
\vspace{-3mm}
\begin{table}[htb]
    \centering
    \vspace{-2mm}
    \caption{Cross-platform evaluation results on SD1.5. PSNR-\uppercase\expandafter{\romannumeral1} \& MSSSIM-\uppercase\expandafter{\romannumeral1} and PSNR-\uppercase\expandafter{\romannumeral2} \& MSSSIM-\uppercase\expandafter{\romannumeral2} measure the average quality of recreated images under two different sample schedulers, respectively.}
    \vspace{-2mm}
    \setlength{\tabcolsep}{2pt}
    \resizebox{0.5\textwidth}{!}{
    \begin{tabular}{cc|cc|cc|cc}
    \toprule
    Saving Device & Time (s) & Recreation Device &  Time (s) & PSNR-\uppercase\expandafter{\romannumeral1} & MSSSIM-\uppercase\expandafter{\romannumeral1} & PSNR-\uppercase\expandafter{\romannumeral2} & MSSSIM-\uppercase\expandafter{\romannumeral2} \\
    \midrule
    CPU~1 & 42.96 & CPU~2 & 82.62 & 80.65 & 1.0000 & 73.68 & 1.0000 \\
    CPU~1 & 42.96 & CPU~3 & 53.41 & 80.20 & 1.0000 & 75.09 & 1.0000 \\
    GPU~1 & 4.71 & GPU~2 & 3.29 & lossless & 1.0000 & 57.90 & 0.9999 \\
    GPU~1 & 4.71 & GPU~3 & 8.61 & 58.44 & 0.9999 & 51.31 & 0.9980 \\
    CPU~1 & 42.96 & GPU~1 & 4.71 & 58.45 & 0.9999 & 51.31 & 0.9980 \\
    CPU~2 & 86.62 & GPU~2 & 3.29 & 58.44 & 0.9999 & 51.24 & 0.9981 \\
    CPU~3 & 53.41 & GPU~3 & 8.61 & 80.10 & 1.0000 & 74.93 & 1.0000 \\
    \bottomrule
    \end{tabular}
    }
    \label{tab:sdv1.5}
\end{table}
\myparagraph{Cross-platform image recreation.} \label{sec:exp_cross_platform}
In the previous experiments, we assume the image saving and recreation processes are conducted on identical software and hardware platforms. However, in real-world applications, the recreation platform often differs from the one described in the image file, resulting in potential information loss. We attribute this primarily to the non-deterministic nature of matrix multiplication across various hardware architectures. Therefore, we evaluated cross-platform image recreation performance, considering three scenarios: cross-CPU, cross-GPU, and the communication between CPU and GPU. The specific hardware models are shown in Table~\ref{tab:notation}. Table 5 presents the cross-platform evaluation results on different CPU and GPU devices.
Due to the excessive time required for generating an image with a resolution of 1024x1024 on CPU platforms (exceeding 10 minutes), we reduce the resolution to 512x512 and the sampling steps to 25 for comparison. We also found that the recreation quality depends on the choice of sampling scheduler, with PSNR-\uppercase\expandafter{\romannumeral1} \& MSSSIM-\uppercase\expandafter{\romannumeral1} corresponding to DDIM sampler~\cite{ddim} and PSNR-\uppercase\expandafter{\romannumeral2} \& MSSSIM-\uppercase\expandafter{\romannumeral2} corresponding to DPM++ 2M sampler~\cite{dpm++}. We observe that images generated across different CPUs and identical GPU models maintain high consistency under various test conditions. Furthermore, DPM++ 2M sampler often negatively impacts cross-platform consistency, likely due to the complex multi-step calculations that amplify differences in floating-point matrix computations. These findings highlight the importance of standardizing hardware for running AI-generated content models. Nevertheless, given that distortions at PSNR values above 50 are imperceptible to the human eye, we consider the inconsistency of platforms only slightly degrades the recreated AIGC images.

\begin{figure}[h]
    \centering
    \begin{subfigure}[b]{0.25\linewidth}
        \includegraphics[width=\linewidth]{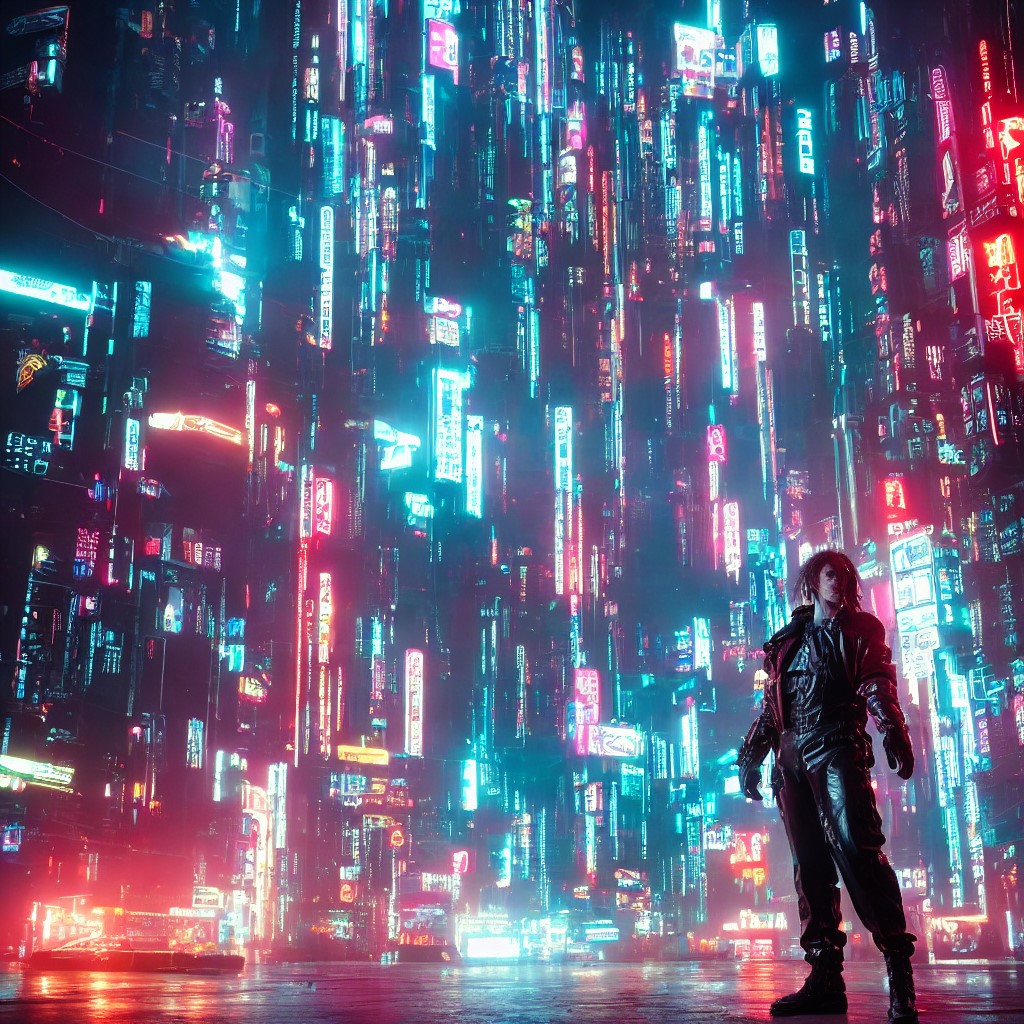}
        \caption{SD1.5}
        \label{fig:SD1.5}
    \end{subfigure}
    \begin{subfigure}[b]{0.25\linewidth}
        \includegraphics[width=\linewidth]{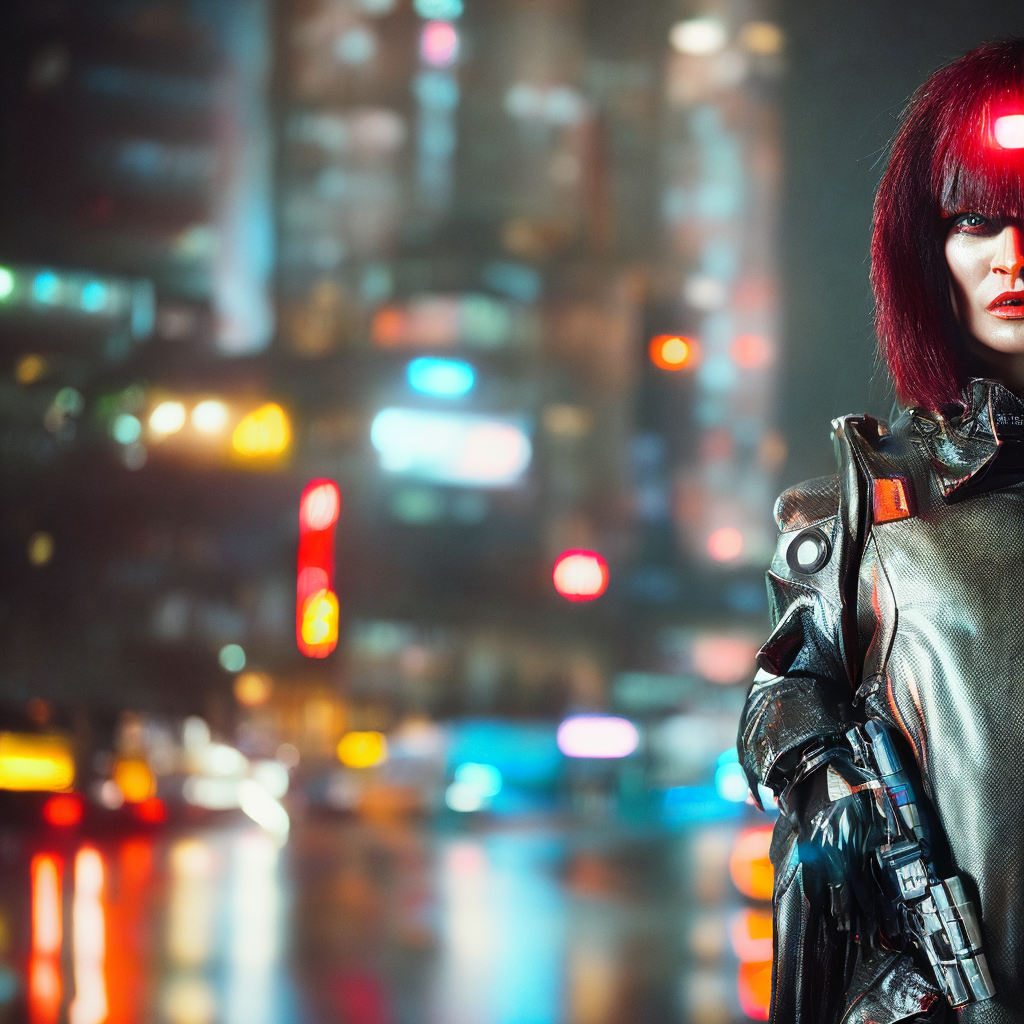}
        \caption{SD2.1}
        \label{fig:image2}
    \end{subfigure}
    \begin{subfigure}[b]{0.25\linewidth}
        \includegraphics[width=\linewidth]{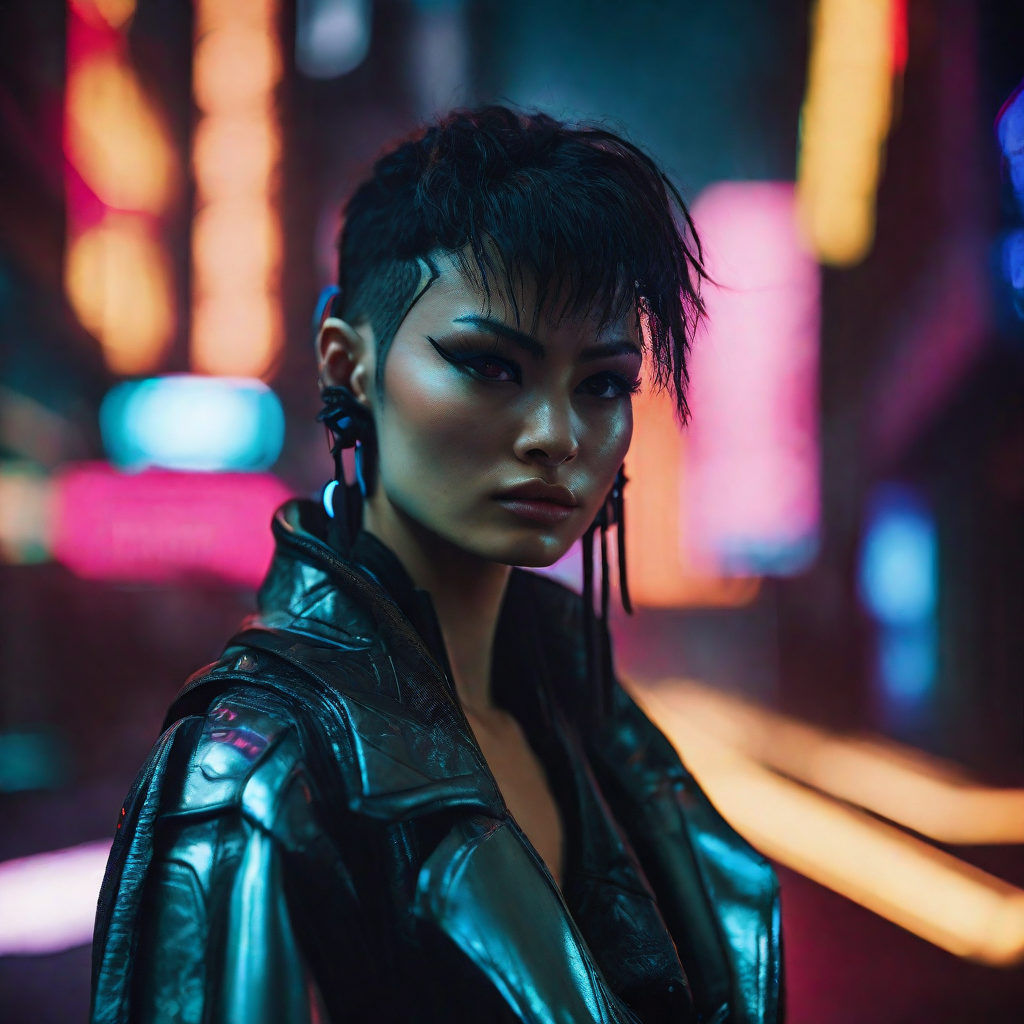}
        \caption{SDXL}
        \label{fig:image3}
    \end{subfigure}
    \vspace{-1mm}
    \caption{Comparison of image generation across different models with identical hyperparameter settings.}
    \label{fig:crossmodel}
\end{figure}

\myparagraph{Cross-model image recreation} \label{sec:exp_cross_model}
Fig.~\ref{fig:crossmodel} demonstrates that when we only change the model configuration and keep the other information unchanged, images generated by different models do not exhibit pixel-level consistency. This underscores the necessity of storing model information in our file format for lossless recreation of AI-generated images.
\vspace{-3mm}

\vspace{-4mm}
\section{Limitations}
\label{sec:limitation}
The potential limitation of our AIGIF is the computational cost on the decoder side, which arises from generation process and is the same as the challenge faced by existing diffusion model-based image compression methods~\cite{yang2024lossy,careil2023towards,hoogeboom2023high,xu2024idempotence,ghouse2023residual,gao2024unimic}. We believe it can be effectively eliminated with the rapid development of efficient image generation methods~\cite{Shang_2023_CVPR,NEURIPS2023_f1ee1cca,huang2024knowledge,Patel_2024_CVPR,castells2024ld}.



\section{Conclusion}
In this paper, we consider a definition of an AI-generated image format, named AIGIF, enabling storing and transmitting high-quality AIGC images at an ultra-low bitrate by compressing the generation syntax rather than pixel data. 
Through systematically investigating the impacts of platform, generative model, and data configurations, we developed a well-defined composable bitstream structure, which achieves ultra-low bitrate compression for AIGC images with a compression ratio of up to 1/10,000, while still ensuring high image fidelity.
To mitigate potential issues related to the loss of generation syntax elements in some specific scenarios, our AIGIF also supports direct compression and transmission of original AIGC images.
Additionally, AIGIF includes an expandable syntax, "exp code," to support new generative models, ensuring adaptability and scalability. 



\bibliographystyle{IEEEtran}
\bibliography{main}

\end{document}